\documentclass[conference]{IEEEtran}
\IEEEoverridecommandlockouts
\usepackage{cite}
\usepackage{comment}
\usepackage{hyperref}
\usepackage{amsmath,amssymb,amsfonts}
\usepackage{algorithmic}
\usepackage{graphicx}
\usepackage{textcomp}
\usepackage[dvipsnames]{xcolor}
\usepackage{dirtree}
\usepackage{soul}
\usepackage{pifont}
\usepackage{tikz}
\usepackage{multirow}
\usepackage{color, colortbl}	
\usepackage{amssymb}
\usepackage{soul}
\usepackage{makecell}

\usepackage{booktabs}
\usepackage{xcolor}
\usepackage{tcolorbox}
\tcbset{colback=white, boxrule=0.4pt, arc=2pt, left=2pt, right=2pt, top=2pt, bottom=2pt}

\definecolor{LightCyan}{rgb}{0.88,1,1}

\usepackage[a4paper, total={184mm,239mm}]{geometry}
\def\BibTeX{{\rm B\kern-.05em{\sc i\kern-.025em b}\kern-.08em
    T\kern-.1667em\lower.7ex\hbox{E}\kern-.125emX}}

\renewcommand{\baselinestretch}{0.975}

\IEEEoverridecommandlockouts
\IEEEpubid{\makebox[\columnwidth]{
979-8-3195-3997-7/26/\$31.00~\copyright2026
IEEE \hfill} \hspace{\columnsep}\makebox[\columnwidth]{ }}

\begin{document}
\bstctlcite{IEEEexample:BSTcontrol}
\title{SAMpLE: A SystemC-AMS Machine LEarning-based Framework for Virtual Prototyping
 \thanks{This work has received funding from the European Chips
Joint Undertaking under Framework Partnership Agreement
No. 101139789 (HAL4SDV).}
}

\author{\IEEEauthorblockN{Andrei Mihai Albu and Sara Vinco}
\IEEEauthorblockA{ 
 Politecnico di Torino, Italy -- Emails: andrei.albu@polito.it, sara.vinco@polito.it}
}

\maketitle

\begin{abstract}
Machine Learning (ML) is increasingly used in virtual prototypes of embedded systems to model behaviors that are difficult to capture analytically. However, integrating ML models into virtual platform simulation is still typically done through ad hoc solutions, which limits reuse, comparability, and reproducibility.
This paper presents \textbf{\textit{SAMpLE}}, an open-source SystemC-AMS-based framework that integrates ML models as first-class Timed Dataflow (TDF) components through a standardized plug-and-play interface. SAMpLE provides two execution backends: a native C++ backend for online training of lightweight models, and an offline backend for executing externally developed models without requiring re-implementation in C++ or manual integration steps.
The framework uses ONNX as a standard model exchange format to enable integration of externally trained ML models into SystemC-AMS simulations, and allows the evaluation of different ML-based solutions within the same testbench, dataset, and simulation workflow. The modular design and unified and reproducible environment will allow future extensions of SAMpLE to new models, without modifying the SystemC-AMS structure.
\end{abstract}

\begin{IEEEkeywords}
SystemC-AMS, Virtual Prototyping, Machine Learning, Timed Dataflow (TDF), Hardware/Software Co-Design.
\end{IEEEkeywords}
\section{Introduction}
\label{sec:introduction}

Machine Learning (ML) techniques are increasingly adopted in embedded-system design to approximate complex behaviours that are difficult to capture with analytical models or traditional simulation approaches, including power consumption, sensor dynamics, ageing effects, and fault prediction \cite{Intro_00, Intro_01, Intro_02}. Compared to conventional modeling techniques, ML-based surrogates provide a favourable trade-off between accuracy, computational cost, and data-driven adaptability, making them increasingly relevant in system-level virtual prototyping.

Despite this trend, virtual platform languages, such as SystemC and its extensions, do not provide a native or standardized mechanism for integrating ML models into simulation semantics~\cite{Intro_04, Intro_05}. As a result, ML integration requires manual integration or on external, ad-hoc extensions layered on top of the simulation infrastructure. Existing approaches are therefore fragmented and tool-specific: ML inference is embedded into custom TLM interfaces targeting fixed application domains~\cite{Intro_08}, delegated to external Python runtimes with strong toolchain coupling~\cite{Intro_06}, or manually re-implemented in C++ at the cost of significant engineering effort and limited model portability. These limitations hinder systematic integration of heterogeneous ML techniques within a unified simulation environment.

This work introduces \textbf{SAMpLE} (\textit{Systemc-Ams Machine LEarning-based framework for virtual prototyping}), a framework that enables direct execution of ML models as native TDF  components. The goal of SAMpLE is to extend SystemC-based virtual platforms with support for ML models that can either \emph{learn online}, adapting incrementally sample-by-sample during simulation execution, or be \emph{trained offline} and loaded into the simulation via ONNX at elaboration time.
 
TDF was chosen as its semantics naturally aligns with ML inference: both operate on causal, sample-driven execution, where fixed-size input vectors are transformed into outputs at discrete activation steps. This equivalence allows ML models to be executed directly within the \texttt{processing()} semantics of TDF, removing the need for co-simulation layers, inter-process communication, or model translation.

SAMpLE implements this integration through a lightweight module abstraction that decouples simulation semantics from inference execution. The framework provides two interchangeable back-ends: 
\begin{enumerate}
    \item \emph{ONLINE backend}: a native C++ implementation designed for light-weight models and online-adaptive learning algorithms, tightly integrated with the SystemC-AMS simulation kernel;
    \item \emph{OFFLINE backend}: an ONNX Runtime-based implementation \cite{ONNX} that enables transparent deployment and load of models trained using heterogeneous ML frameworks, e.g., PyTorch, TensorFlow, and scikit-learn \cite{PyTorch,TensorFlow,Scikit-learn}.
\end{enumerate}
Both backends conform to a unified interface, enabling heterogeneous ML models to coexist and run under identical SystemC-AMS simulation conditions.

The main contributions of this work are thus as follows:
\begin{itemize}
    \item A native integration methodology that \emph{embeds ML models directly within SystemC-AMS TDF semantics}, eliminating the need for external co-simulation frameworks, middleware layers, or manual model translation;
    \item A \emph{unified \texttt{MLModule} abstraction} that separates simulation behavior from inference execution, enabling interchangeable ML backends within a single simulation environment;
    \item A \emph{dual-backend architecture} supporting both lightweight native C++ implementations and portable ONNX-based deployment of models trained using heterogeneous ML frameworks;
    \item An \emph{open-source reproducible evaluation framework} and Language Reference Manual (LRM)-compliant lifecycle management infrastructure supporting deterministic dataset handling, CSV-driven experimentation, and cross-framework model parity verification \cite{sample_tool}. 
\end{itemize}

\section{Background and Related Works}
\label{sec:background}

\subsection{SystemC and SystemC-AMS}
{SystemC} is a C++-based system-level modelling framework standardized under IEEE 1666 and widely adopted for the design and simulation of complex embedded and heterogeneous systems \cite{SystemC}. By extending standard C++ with HW-oriented abstractions, SystemC enables the unified modelling of HW and SW components within a common simulation environment.

To support mixed-signal and cyber-physical systems, the {SystemC Analog/Mixed-Signal (SystemC-AMS)} extension introduces additional Models of Computation (MoCs) tailored to analogue and dataflow-oriented simulations \cite{SystemC_AMS}, and is suited for multi-domain virtual platform development \cite{Related_04, Related_05, Related_06}. Among the available MoCs, the TDF represents systems as interconnected processing blocks that periodically consume and produce samples according to statically scheduled execution semantics. Modules communicate through ports and operate at predefined rates and timesteps, enabling deterministic, causal, and computationally efficient simulation while preserving temporal behavior.

\subsection{ONNX}

Open Neural Network Exchange (ONNX) is an open standard for representing machine learning and deep learning models in a platform-independent format \cite{ONNX}. Originally introduced by Microsoft and Meta, ONNX is designed to improve interoperability among machine learning frameworks, compilers, and hardware accelerators. The format enables models trained in one framework (e.g., PyTorch \cite{PyTorch} or TensorFlow \cite{TensorFlow}) to be executed across different inference engines and hardware platforms without requiring substantial modifications.

In this work, ONNX models are executed using the ONNX Runtime library \cite{onnxruntime}, which provides a high-performance inference engine with support for graph optimizations and multiple execution backends (e.g., CPU and hardware accelerators). This runtime layer serves as the bridge between the exported model graph and the SystemC-AMS simulation environment, enabling deterministic evaluation of machine learning inference within the TDF execution flow.

\subsection{Related Work}
\label{subsec: related-works}

The integration of ML techniques into system-level design environments has received increasing attention due to the growing adoption of data-driven methodologies in embedded and cyber-physical systems \cite{Intro_03,Related_01}. ML models are increasingly employed to approximate computationally intensive behaviors, including power estimation, sensor modeling, fault prediction, adaptive control, workload characterization, and reduced-order system representations \cite{Related_02, Related_03}. As a consequence, virtual prototyping environments are progressively required to support the execution of heterogeneous ML workloads alongside traditional system-level simulation models.

SystemC and SystemC-AMS are widely adopted for Electronic System-Level (ESL) design, virtual prototyping, and mixed hardware/software simulation \cite{Related_04, Related_05, Related_06}. Several methodologies have extended SystemC-based environments to support increasingly complex embedded workloads, including distributed simulation, heterogeneous system modeling, and transaction-level virtual platforms \cite{Related_07, Intro_04}. More recently, SystemC has also been employed to model and evaluate AI accelerators and deep-learning-oriented architectures during early design-space exploration \cite{Related_08, Related_09}.

Despite these advances, current SystemC-based methodologies do not provide a standardized mechanism for integrating native ML models directly as native simulation components. Existing approaches typically rely on one of three strategies: first, export models as Functional Mock-up Units (FMUs) according to the Functional Mock-up Interface (FMI) standard, enabling their execution within heterogeneous co-simulation environments. In such setups, ML models and system-level components are instantiated as independent FMUs and orchestrated by a master algorithm, allowing joint simulation across different tools. However, while this improves interoperability, it still introduces non-negligible synchronization overhead \cite{Intro_06,Related_10,Related_11, FMI_reference}.  Second, trained models are frequently reimplemented manually in C++ to enable tight integration within the simulation kernel\cite{Related_08}. Third, application-specific interfaces are developed around particular ML libraries or accelerator frameworks \cite{Intro_08,Related_12}. Overall, although effective for specific use cases, these solutions generally reduce portability, increase maintenance complexity, and complicate reproducibility across simulation environments.

\section{Methodology}
\label{sec:methodology}

This section presents the SAMpLE methodology, outlined in Figure \ref{fig:SAMpLE_architecture_overview}.
Subsection~\ref{subsec:overview} introduces the architecture, that is detailed in the following Subsections B-D, while Subsection~\ref{subsubsec:standard} discusses current limitations and outlines a proposal for standardisation.
SAMpLE is released open-source at \cite{sample_tool}.

\subsection{Architecture Overview}
\label{subsec:overview}

\begin{figure*}[t]
    \centering
    \includegraphics[width=\linewidth]{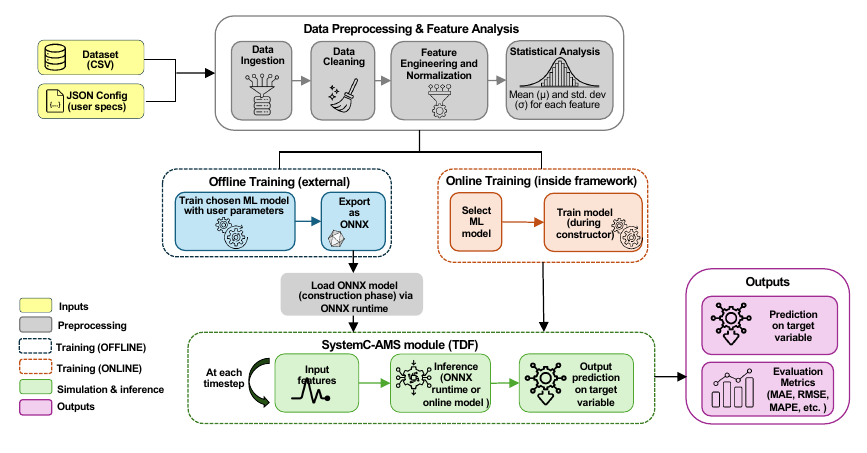}
    \vspace{-1cm}
    \caption{Architecture overview of the SAMpLE framework.  The
      workflow is driven by a dataset and a JSON-based configuration
      file.  The preprocessing stage produces a standardised dataset
      representation and normalisation parameters.  Two execution
      modes are supported: offline training through ONNX-exported
      models loaded via ONNX Runtime, and online training performed
      directly inside the SystemC-AMS module.  During Timed Dataflow
      (TDF) simulation, the selected backend performs inference at
      each timestep, producing predictions and runtime evaluation
      metrics (MAE, RMSE, $R^{2}$).}\vspace{-0.3cm}
    \label{fig:SAMpLE_architecture_overview}
\end{figure*}

The objective of SAMpLE is to provide a unified SystemC-AMS framework
for integrating ML models into the early design evaluation of virtual
platforms, enabling reproducible and comparable ML-driven assessment
across heterogeneous model families. As shown in
Fig.~\ref{fig:SAMpLE_architecture_overview}, the workflow is driven by
two inputs: a CSV dataset describing the system behaviour to be
modelled and a JSON configuration file (\texttt{config.json}) that
serves as the single source of truth for all subsequent stages.

The process begins with dataset ingestion and preprocessing. Then, depending on the selected configuration, models are either trained offline in a Python environment and imported into the SystemC-AMS module via ONNX Runtime or trained directly within the framework during the 
initialisation phase. In both cases the model is exposed to the simulation through a single \texttt{IMLModel} interface, so that the TDF scheduler interacts with every model family in an identical manner. This decoupling of \emph{simulation semantics} from \emph{inference execution} is the property that the rest of the methodology builds upon.

\subsection{Data Preprocessing and Feature Analysis}
\label{subsec:preprocessing}

The preprocessing stage is implemented as a modular, configuration-driven pipeline that promotes reproducibility and portability across different simulation backends. All parameters, including file paths, feature schemas, and processing options, are specified in a centralized JSON configuration file, thereby eliminating hard-coded dependencies. The \texttt{config.json} exposes 18 keys across five sections (model, dataset, simulation, validation, debug), all optional with framework defaults.

\texttt{dataset.csv} requires one row per timestep, a target column, and named feature columns (inline or via file), with no other structural constraints. The pipeline comprises four stages. First, \emph{data ingestion} loads the raw dataset and extracts timestep information. Second, \emph{data cleaning} removes missing or invalid entries. Third, \emph{feature engineering} derives temporal and aggregate features and applies cyclical encoding to periodic temporal attributes to preserve their inherent periodicity (e.g., hour-of-day or day-of-week). Cyclical encoding eliminates artificial discontinuities at period boundaries (e.g., between hours~23 and~0) by mapping each periodic feature to a two-dimensional representation:

\begin{equation}
    x_{\sin} = \sin\!\left(\frac{2\pi\, t}{T}\right) \quad
    x_{\cos} = \cos\!\left(\frac{2\pi\, t}{T}\right)
\end{equation}
\noindent
where $t$ is the raw feature value and $T$ is the period defined in
the configuration file (e.g.\ $T{=}24$ for hourly, $T{=}7$ for
weekly cycles).  This design ensures consistency across datasets
without requiring source-code modifications.

The fourth stage performs \emph{statistical analysis}, computing descriptive metrics such as the mean, standard deviation, minimum, and maximum values of each feature. These statistics are subsequently used both as auxiliary inputs for machine-learning workflows and as reference values during model evaluation. To support dataset inspection and validation, the framework also provides a visualization utility that generates plots of the raw data together with the computed statistical summaries.

Finally, the processed dataset is serialized into \texttt{.csv} format and partitioned into training, validation, and test subsets using a seed-controlled, chronology-preserving procedure. The test partition is subsequently consumed during SystemC-AMS simulation to evaluate the predictive performance of the generated machine-learning models under execution conditions representative of the target system.

\subsection{Training Phase}
\label{subsec:training}

SAMpLE supports two complementary training paradigms that share the same preprocessing pipeline and simulation interface, differing only  in when model fitting is performed: \emph{offline training} path, that allows the user to load  already existing models or to train the model with a user-defined pipeline (Section~\ref{subsubsec:offline}), and an \emph{online training} path, where all training is handled by SAMpLE, with few user-defined parameters Section~\ref{subsubsec:online}.

\subsubsection{Online training}
\label{subsubsec:online}

In the online training path, model construction and parameter initialization are performed automatically at simulation start-up, eliminating the need for a separate training phase. The framework provides a set of predefined training strategies that can be selected through the configuration file, enabling rapid deployment of predictive models with minimal manual intervention. Although this
approach may not achieve the accuracy attainable through extensive offline hyperparameter optimization, it significantly reduces model development effort and accelerates exploration of design alternatives.

At the core of the online mechanism is a unified predictor
abstraction, \texttt{IMLModel}, that separates feature ingestion,
prediction, and adaptation logic. Each concrete predictor implements
four functions:
\begin{itemize}
    \item \texttt{required\_window\_size()}: allows to set how
  many past samples the model needs before its first valid output;
    \item \texttt{predict(window)}: maps a fixed-size
  sliding window of past feature vectors to one or more output
  predictions;
    \item \texttt{update(target)}: performs an incremental
  parameter step against the most recent ground-truth observation;
 \item \texttt{supports\_training()}: exposes at runtime
  whether the model supports in-framework adaptation.
\end{itemize}

The current version of SAMpLE supports a number of ML model types, including adaptive linear estimators and nonlinear or regime-aware learning approaches, providing a broad basis for modelling time-varying dynamic systems. This set will be extended as part of future work. The type of model to be constructed, together with any parameters, can be set by the user in the initial configuration file. 

The supported models include adaptive linear, ensemble-based, regime-switching, and nonlinear online learning approaches. NLMS-ARX and EW-RLS-ARX extend ARX models with recursive adaptation mechanisms, respectively based on normalized least-mean-squares and exponentially weighted recursive least squares \cite{nlms_arx,7402726}. The Hedge Ensemble dynamically combines multiple predictors through online weight updates \cite{arora2012multiplicative}, while GHMM-Regime captures latent operating regimes using Gaussian hidden Markov models \cite{wang2020regime}. Nonlinear adaptation is addressed by EKF-MLP, which trains multilayer perceptrons through extended Kalman filtering \cite{gaamouri2018denoising}, and by KRLS-ALD, which performs sparse kernel recursive least-squares regression using an approximate linear dependency criterion \cite{liu2022sparse}. Overall, these models provide complementary capabilities in terms of interpretability, adaptivity, nonlinear modelling power, and computational efficiency.

\subsubsection{Offline training}
\label{subsubsec:offline}

In the offline training path, model fitting is performed outside the SAMpLE framework using arbitrary machine-learning workflows and toolchains. This approach supports two deployment scenarios: (i) training custom models using the preprocessed datasets generated by SAMpLE, and (ii) importing previously trained models without requiring execution of the preprocessing and training stages. As a result, existing machine-learning assets can be reused directly within the
simulation environment.

To support framework-independent deployment, SAMpLE adopts the Open Neural Network Exchange (ONNX) format as a common model representation. Models trained using external frameworks are exported to ONNX and wrapped to conform to the \texttt{IMLModel} interface (e.g., \texttt{required\_window\_size()} and \texttt{predict(window)}). The resulting ONNX model, together with the serialized normalization and scaling parameters, constitutes a self-contained deployment artifact.

The ML model is loaded at simulation initialization time, and required two validation steps: 
\begin{itemize}
    \item a static input-shape check, verifying that the number
of ONNX input nodes matches the configured feature-vector width; 
\item a dry-run inference on a zero-padded vector confirming
that the graph is callable and produces a finite output.
\end{itemize}

These guards ensure that the simulation-time environment faithfully
replicates the Python training environment without embedding an
interpreter or requiring inter-process communication.

\subsection{Simulation Semantics}
\label{subsec:simulation}

From a semantic perspective, ML inference aligns naturally with the TDF execution model: given a fixed set of input features sampled at regular intervals, inference produces a deterministic output at each activation; a behavior that maps directly onto the causal, periodically-scheduled signal processing abstraction that TDF modules implement. This correspondence suggests that trained ML models can be represented as TDF processing blocks, without requiring auxiliary co-simulation layers or external execution infrastructures.

SAMpLE leverages this correspondence by integrating ML model construction, deployment, and inference directly within the TDF execution flow. Unlike co-simulation approaches or manually translated implementations, the proposed framework introduces a unified abstraction layer that preserves the semantics of the underlying TDF model while supporting multiple machine-learning backends. This design ensures deterministic execution, reproducibility, and portability
across simulation environments.

\begin{figure}[!h]
    \centering
    \includegraphics[width=\linewidth]{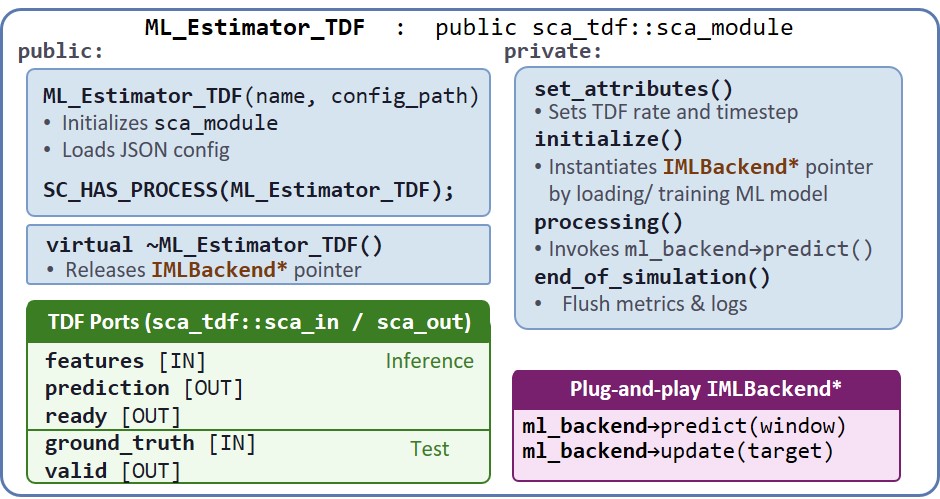}
    \caption{Structure of the proposed SystemC-AMS primitive. It preserves the TDF lifecycle, features an instance of \texttt{IMLBackend} to load and handle the ML model, and features a fixed interface with mandatory inference ports and optional test ports.} 
    \label{fig:sca-ml-module}
\end{figure}

Figure~\ref{fig:sca-ml-module} illustrates the structure of the proposed TDF primitive. Input features are received through the \texttt{features} port, while predictions are emitted through the \texttt{prediction} port. A dedicated \texttt{ready} signal indicates when sufficient historical samples have been accumulated to produce valid predictions. Two optional ports, \texttt{ground\_truth} and \texttt{valid}, support online evaluation and validation of model outputs during simulation.

The proposed primitive fully complies with the standard TDF execution
semantics and lifecycle:

\paragraph{Simulation setup}
The constructor allows to instantiate the module and loads the information contained in the input configuration file. The \texttt{set\_attribute} primitive extrapolates such information to set the module time step and the update rate of output ports. The \texttt{initialize} function instantiates an \texttt{IMLBackend} object with two different behaviors, depending on the type of training:
\begin{itemize}
    \item \emph{offline training}: loads the ML model trained offline via ONNX, sets up its parameters based on the initial configuration, and triggers the validity checks (in case of failure, simulation is stopped via \texttt{sc\_stop()});
    
    \item \emph{online training}: the chosen ML model is built starting from the initial configurations, the statistical information derived from the feature analysis, and the training set.
\end{itemize}

\paragraph{Simulation and inference}
The \texttt{processing} function triggers one inference of the ML model (e.g., via invocation of the \texttt{predict} function) and updates the \texttt{prediction} port at any activation time step.
Given that the model may require a window size to achieve warmup, the \texttt{ready} is set to true only after \texttt{required\_window\_size()} ticks, to notify that the generated value can now be used as valid data. The current implementation supports single-step-ahead prediction, although the abstraction can be extended to multi-step forecasting strategies in future work.

\paragraph{Model test and validation} When the optional \texttt{ground\_truth} signal is asserted, the processing stage additionally evaluates prediction accuracy against reference values from the test set. Error statistics are accumulated throughout the simulation, and the \texttt{valid} output is asserted
whenever a prediction violates the configured acceptance criteria. At
simulation termination, the framework reports standard regression
metrics, including coefficient of determination ($R^2$), Root Mean Squared Error (RMSE), and Mean Absolute Error (MAE).

\paragraph{Backend interchangeability}
The definition of a common \texttt{IMLModel} interface for both online and offline trained models and of a standard interface allows to make 
e.g. an
ONNX-exported Random Forest externally indistinguishable from one
containing a native online predictor. %
This interchangeability allows to compare different model
families (e.g., classical estimators, tree ensembles, deep sequence
models) on the same testbench with the same
dataset split and metric code, without model-specific evaluation
infrastructure. Additionally, the simulation-level cost of a new model
(inference latency, memory footprint) is measured directly inside the
virtual platform rather than in an isolated benchmark.

\subsection{Toward a standardised SystemC-AMS primitive}
\label{subsubsec:standard}

The integration patterns established by SAMpLE suggest that the definition of a similar primitive in the SystemC-AMS standard would be beneficial.  ML surrogates are now routine components of production virtual platforms, yet the standard provides no mechanism to host them, and the only sample generation primitive remains the sinusoidal signal source \texttt{sca\:lsf::sca\_source}. On the other hand, the \texttt{IMLModel} contract (windowed input, fixed-size prediction output, optional ground-truth update, \texttt{ready} flag) has proved sufficient to integrate a number of heterogeneous models, from a compact NLMS-ARX to more complex ones. 
Finally, the trade-off between ecosystem portability (ONNX) and per-tick
adaptability (native C\texttt{++}) is structural; the dual-backend
design will remain necessary regardless of how individual ML
frameworks evolve.
The proposal is intentionally conservative: it adds one module type with no additional impact on the language, and avoids tool-vendor
lock-in by shipping two interchangeable reference backends.

\section{Experimental Evaluation}
\label{sec:results}

We evaluate \textsc{SAMpLE} with respect to three directions:
\begin{itemize}
  \item[\textbf{Q1}] \textit{Fidelity:} does SystemC-AMS execution reproduce the numerical behaviour of Python models after ONNX export?
  \item[\textbf{Q2}] \textit{Heterogeneity:} can structurally diverse ML models be executed through a unified \texttt{IMLModel} interface without modifying the SystemC netlist?
  \item[\textbf{Q3}] \textit{Portability:} does the pipeline transfer across datasets within the same application domain without changes to the simulator source code?
\end{itemize}

The goal of the evaluation is not to optimise individual ML models, which can be tuned independently via the \texttt{config.json} file, but to rather assess the generality and interoperability of the proposed framework across heterogeneous model classes and datasets.

\subsection{Experiments setup}
We consider two datasets with distinct temporal characteristics. {UCI Appliances}~\cite{appliances_energy_prediction_374} contains 19,735 samples at 10-minute resolution and represents noisy, occupancy-driven residential energy consumption ranging from 10 to 1,080Wh. {Tetuan City}~\cite{power_consumption_of_tetouan_city_849} contains 52,416 samples at the same resolution and represents smooth, periodic urban electricity demand between 13.9 and 52.2kW. Both datasets are split chronologically to avoid temporal leakage. 

For offline training, we supported ARX Ridge, which applies ridge-regularized least squares to ARX regressors to improve robustness under correlated inputs \cite{kibria2019ridge}, Random Forest, which performs nonlinear prediction through an ensemble of randomized decision trees \cite{dudek2022comprehensive}, and XGBoost, which uses regularized gradient-boosted decision trees for scalable and accurate regression \cite{chen2016xgboost}.

All experiments run on an Intel Core i7-10700 with 16\,GB RAM under Ubuntu 22.04, and all artifacts (configurations, datasets, models, predictions, and metric reports) are released alongside the source code for exact reproducibility \cite{sample_tool}. 

For each experiment, we report total wall-clock simulation time\footnote{Time: offline reports Python CPU training time; simulation fidelity reports TDF validation time; online reports wall-clock simulation time, including elaboration and scheduling.}, $R^2$, RMSE, and MAE, expressed in Wh for UCI, W for Tetuan.

\begin{table*}[th]
\centering
\caption{Offline training and simulation fidelity results.
         Chronological 15\,\% test split (2\,954 UCI / 7\,855 Tetuan
         samples). Time denotes CPU training time for offline models and
         SystemC-AMS TDF validation time for simulation fidelity experiments.
         RMSE and MAE are reported in Wh (UCI) and W (Tetuan).}
\label{tab:offline-sim}
\setlength{\tabcolsep}{3pt}
\renewcommand{\arraystretch}{0.95}
\begin{tabular}{l
  >{\centering\arraybackslash}p{3.8em}
  >{\centering\arraybackslash}p{4.8em}
  >{\centering\arraybackslash}p{4.8em}
  >{\centering\arraybackslash}p{5.0em}
  >{\centering\arraybackslash}p{3.8em}
  >{\centering\arraybackslash}p{4.8em}
  >{\centering\arraybackslash}p{4.8em}
  >{\centering\arraybackslash}p{5.0em}
}
\toprule
& \multicolumn{4}{c}{\textbf{UCI Appliances}}
& \multicolumn{4}{c}{\textbf{Tetuan City}} \\
\cmidrule(lr){2-5}\cmidrule(lr){6-9}
\textbf{Model}
  & $\mathbf{R^{2}}$ & \textbf{RMSE} & \textbf{MAE} & \textbf{Time (s)}
  & $\mathbf{R^{2}}$ & \textbf{RMSE} & \textbf{MAE} & \textbf{Time (s)} \\
\midrule
\multicolumn{9}{l}{\textit{Offline training (Python)}} \\[1pt]
ARX  & \textbf{0.278} & \textbf{77.3} & \textbf{42.9} & 0.020
     & \textbf{0.986} & \textbf{721.6} & \textbf{503.4} & 0.600 \\
XGB  & 0.235          & 79.6          & 48.4          & 0.100
     & 0.985          & 755.4         & 535.9         & 2.900 \\
RF   & 0.148          & 83.9          & 59.3          & 0.600
     & 0.974          & 984.1         & 708.7         & 8.100 \\
\midrule
\multicolumn{9}{l}{\textit{Simulation fidelity (SystemC-AMS TDF)}} \\[1pt]
ARX  & 0.278          & 77.3          & 42.9          & 0.206
     & 0.986          & 721.6         & 503.9         & 0.343 \\
XGB  & 0.238          & 79.8          & 48.4          & 0.209
     & 0.985          & 755.4         & 535.9         & 0.350 \\
RF   & 0.148          & 83.9          & 59.3          & 0.262
     & 0.974          & 984.1         & 708.7         & 1.478 \\
\bottomrule
\multicolumn{9}{l}{\vspace{0.5em}\footnotesize
  \textbf{Bold}: best per metric per dataset.}
\end{tabular}
\vspace{-0.5cm}
\end{table*}

\begin{table*}[th]
\centering
\caption{Online adaptive forecasting results. Chronological 20\,\%
         test split (3\,947 UCI / 10\,484 Tetuan samples). Time denotes
         wall-clock SystemC-AMS simulation time, including model execution,
         initialization overhead, and TDF scheduling. RMSE and MAE are reported
         in Wh (UCI) and W (Tetuan).}
\label{tab:online}
\setlength{\tabcolsep}{3pt}
\renewcommand{\arraystretch}{0.95}
\begin{tabular}{l
  >{\centering\arraybackslash}p{3.8em}
  >{\centering\arraybackslash}p{4.8em}
  >{\centering\arraybackslash}p{4.8em}
  >{\centering\arraybackslash}p{5.0em}
  >{\centering\arraybackslash}p{3.8em}
  >{\centering\arraybackslash}p{4.8em}
  >{\centering\arraybackslash}p{4.8em}
  >{\centering\arraybackslash}p{5.0em}
}
\toprule
& \multicolumn{4}{c}{\textbf{UCI Appliances}}
& \multicolumn{4}{c}{\textbf{Tetuan City}} \\
\cmidrule(lr){2-5}\cmidrule(lr){6-9}
\textbf{Model}
  & $\mathbf{R^{2}}$ & \textbf{RMSE} & \textbf{MAE} & \textbf{Time (s)}
  & $\mathbf{R^{2}}$ & \textbf{RMSE} & \textbf{MAE} & \textbf{Time (s)} \\
\midrule
Hedge Ensemble & \textbf{0.879} & \textbf{31.5} & \textbf{7.0}  & 0.009
               & 0.992  & 550.12   & 369.71   & 0.022    \\
NLMS-ARX       & 0.763          & 43.2          & 20.8          & 0.009
               & \textbf{0.9961}& \textbf{383.2}& \textbf{267.7}& 0.048     \\
Online-GP      & 0.741          & 45.2          & 21.7          & 0.768
               & 0.9858         & 736.4         & 484.7         & 3.602 \\
EW-RLS-ARX     & 0.648          & 52.6          & 17.1          & 0.010
               & 0.9954         & 416.9         & 298.9         & 0.049     \\
GHMM           & 0.595          & 56.0          & 23.8          & 0.009
               & 0.9960         & 391.7         & 285.8         & 0.046     \\
GBT            & 0.565          & 58.5          & 28.2          & 0.009
               & 0.9326         & 1\,601.8      & 1\,322.6      & 0.045     \\
EKF-MLP        & 0.472          & 64.4          & 32.9          & 0.081
               & 0.9689         & 1\,087.7      & 771.1         & 0.313    \\
KRLS-ALD       & 0.268          & 75.9          & 38.1          & 0.045
               & 0.8883         & 2\,062.1      & 1\,578.6      & 0.222    \\
\bottomrule
\multicolumn{9}{l}{\footnotesize
  \textbf{Bold}: best confirmed result per dataset.}\vspace{-0.3cm}
\end{tabular}
\end{table*}

\subsection{Results and Discussion}
\label{subsec:fidelity}

Tables~\ref{tab:offline-sim} and~\ref{tab:online} jointly answer the three research questions. 

For the stateless offline models, \textsc{SAMpLE} reproduces Python predictions (\textbf{Q1)}. Across both datasets, $R^2$, RMSE, and MAE for ARX, XGB, and RF are virtually identical to the Python baseline, with the largest discrepancy being $\Delta R^2 = 0.003$ for XGB on the UCI dataset, within parity constraints. These results indicate that ONNX deployment preserves predictive fidelity with minimal overhead: all offline runs complete in under $1.5\text{s}$, even for ensemble models.

The framework unifies structurally heterogeneous model families under a single SystemC-AMS netlist (\textbf{Q2)}. Three offline backends (linear autoregressive, gradient-boosted trees, and random forest) and eight online backends (adaptive filters, kernel methods, Gaussian processes, regime-switching HMMs, EKF-MLP, KRLS-ALD, Online-GBT, and Online-GP) execute within the same simulation structure, where model switching requires no manual code modification (\textbf{Q3)}. Offline models are selected via a single field in \texttt{config.json}, while online models are instantiated through a compile-time \texttt{IMLModel} dispatch. The same mechanism applies across datasets: transitioning from UCI to Tetuan required only swapping the input CSV and feature schema, with no changes to C++ code, SystemC modules, or build configuration.
The differing predictive behavior across datasets reflects workload complexity rather than implementation error. On the smooth, periodic Tetuan signal, even linear ARX achieves $R^2 = 0.986$, with all offline models clustering closely. In contrast, on the highly non-stationary UCI occupancy trace, offline models degrade substantially to $R^2 \in [0.13, 0.28]$, as their fixed parameters cannot track regime shifts.

This gap is bridged by adaptive online learners natively embedded in the same timed-dataflow execution model. On UCI, the best-performing online method (Hedge Ensemble) reaches $R^2 = 0.879$, more than a threefold improvement over the best offline baseline, due to per-sample adaptation to occupancy-driven dynamics. Other online methods (NLMS-ARX, EW-RLS-ARX, GHMM, Online-GP) occupy a $5$–$30$ percentage-point performance band, while KRLS-ALD and Online-GBT lag, reflecting differences in how well each algorithm matches non-stationary behavior under identical scheduling and metric conditions.

On Tetuan, both offline and online approaches converge, with NLMS-ARX reaching $R^2 = 0.9961$ compared to $0.986$ for offline ARX. In this regime, accuracy differences are secondary to computational cost. NLMS-ARX completes inference in $48\text{ms}$ end-to-end with no external training stage, whereas the offline pipeline requires $0.6\text{s}$ of Python training plus $0.343\text{s}$ of simulation and ONNX export overhead. Across online methods, runtime spans three orders of magnitude, from $9\text{ms}$ (linear filters) to $3.6\text{s}$ (Online-GP on Tetuan), all measured directly within the virtual platform under TDF semantics. This provides the primary contribution of \textsc{SAMpLE}: a unified, in-situ benchmark enabling fair comparison of heterogeneous learning paradigms under identical execution and timing constraints.

\section{Conclusion}
\label{sec:conclusion}
 
This paper presented SAMpLE, an open-source SystemC-AMS framework that integrates ML inference and online learning as native TDF citizens, removing the ad-hoc integration layer that fragments ML-enabled virtual prototyping.

Experimental results show that SAMpLE's in-simulation predictions closely match Python-based evaluations, with only minor cross-environment discrepancies, confirming the effectiveness of the ONNX parity guarantees enforced at construction time. Through the unified \texttt{IMLModel} abstraction, heterogeneous model families are compared under identical SystemC-AMS scheduling, dataset partitions, and metric pipelines.

Future work will extend SAMpLE with heterogeneous model ensembles, uncertainty-aware inference through confidence-aware TDF interfaces, and validation on larger virtual platforms for design space exploration.

\newpage
\bibliographystyle{IEEEtran}

\bibliography{references}

\end{document}